\documentclass{article}

\usepackage[preprint]{neurips_2021}

\usepackage[utf8]{inputenc} 
\usepackage[T1]{fontenc}    
\usepackage{hyperref}       
\usepackage{url}            
\usepackage{booktabs}       
\usepackage{amsfonts}       
\usepackage{nicefrac}       
\usepackage{microtype}      
\usepackage{xcolor}         
\usepackage{graphics} 
\usepackage{times}  
\usepackage{helvet}  
\usepackage{courier}  
\usepackage{natbib}
\usepackage{graphicx}\graphicspath{ {./figures/} }
\usepackage{amsmath} 
\usepackage{amssymb}  
\usepackage{bm}
\usepackage{multicol}
\usepackage{booktabs}
\usepackage{wrapfig}
\usepackage{xcolor}

\usepackage{caption}
\usepackage{subcaption}
\usepackage{multirow}
\usepackage{mathtools}
\usepackage{amsthm}
\usepackage{array}
\usepackage[para]{footmisc}
\usepackage{cite}
\usepackage{istgame}
\usepackage{subcaption}
\usepackage{amsmath}
\usepackage{ upgreek }
\usepackage{multirow}
\usepackage{textcomp}
\usepackage{siunitx}
\usepackage{adjustbox}
\usepackage{bm}

\DeclarePairedDelimiter\floor{\lfloor}{\rfloor}

\title{A taxonomy of strategic human interactions\\ in traffic conflicts}

\author{
    Atrisha Sarkar \textsuperscript{\rm 1},
    Kate Larson \textsuperscript{\rm 1},
    Krzysztof Czarnecki \textsuperscript{\rm 2}
    \\

\textsuperscript{\rm 1}David R Cheriton School of Computer Science\\
\textsuperscript{\rm 2}Department of Electrical and Computer Engineering\\
University of Waterloo, Ontario, Canada\\

    atrisha.sarkar, kate.larson, k2czarne@uwaterloo.ca

}
\begin{document}

\maketitle

\begin{abstract}
In order to enable autonomous vehicles (AV) to navigate busy traffic situations, in recent years there has been a focus on game-theoretic models for strategic behavior planning in AVs. However, a lack of common taxonomy impedes a broader understanding of the strategies the models generate as well as the development of safety specification to identity what strategies are safe for an AV to execute. Based on common patterns of interaction in traffic conflicts, we develop a taxonomy for strategic interactions along the dimensions of agents' initial response to right-of-way rules and subsequent response to other agents' behavior. Furthermore, we demonstrate a process of automatic mapping of strategies generated by a strategic planner to the categories in the taxonomy, and based on vehicle-vehicle and vehicle-pedestrian interaction simulation, we evaluate two popular solution concepts used in strategic planning in AVs, QLk and Subgame perfect $\epsilon$-Nash Equilibrium, with respect to those categories.
\end{abstract}

\section{Introduction}

Human co-ordination and co-operation is central to resolving conflicts in any busy traffic situation. With self-driving cars thrown into the mix, it is clear that behavior planning algorithms for autonomous vehicles (AVs) need to understand and act in a manner that ensures safe co-existence with other human road users for the foreseeable future. Co-ordination among road users is often mediated by traffic rules; however, anyone who has ever walked through a busy city intersection recognizes that humans do not always act in accordance with the prescribed rules, and \emph{ad-hoc} strategic interactions takes the place of strict adherence to rules. In order to equip AVs to participate in such interactions, in recent years there has been a focus on strategic models for AVs, where a set of road users are modelled as players in a general sum game, and various solution concepts have been applied to address the problem of planning \citep{fisac2019hierarchical, tian2018adaptive, li2018game, sadigh2018planning}, as well as the problem of modelling naturalistic human driving behavior \citep{sarkar2021solution, sun2020game, Geiger_Straehle_2021}. The expression of strategies generated by these models are closely tied to the choice of the action space in the game formalism, and covers a wide array of examples, such as specific control actions (acceleration, target velocity, etc.) \citep{li2019decision}, continuous trajectories \citep{Geiger_Straehle_2021}, combination of hierarchical long-horizon and short horizon control actions \citep{fisac2019hierarchical}, and combination of hierarchical high-level maneuver and trajectory patterns \citep{sarkar2021solution}. While these models collectively provide a rich landscape of strategic models for engineers to chose from in the development of AVs, when it comes to a broader understanding, there is a lack of a common language to communicate to all stakeholders, including engineers, regulators, and the broader public, what does these strategic behavior in a given traffic situation on the part of an AV entails. Therefore, there is a need for a taxonomy of strategic interactions that encapsulates common patterns of behavior observed in traffic, as well as a framework for categorizing the strategies generated by the strategic planners into that taxonomy. \par
The need for a taxonomy of strategic interactions is also relevant when it comes to safety analysis of AVs. What is considered as a safe action for an AV ideally should depend on verifiable safety specifications, and frameworks such as Responsibility-sensitive safety (RSS) \citep{shalev2017formal} and Safety Force Field \citep{nister2019safety} provide such frameworks of safety requirements. However, the safety requirements in the frameworks focus on short-term reactive safety (e.g. over instantaneous velocities of vehicles), whereas strategic planner generate a plan over a longer horizon interactions. Since unsafe behavior at a strategic level can lead to hazardous situations eventually, there is a gap in the current safety specification frameworks to address strategic safety. Development of a taxonomy for strategic interactions is a first step towards addressing that gap, thereby allowing for identification of hazardous behaviors of a subject AV in strategic interactions in traffic conflicts. \par
In this paper, we focus specifically on strategic interactions based on modalities of non-verbal communication, e.g., as expressed through kinematic movements. Within the scope of a more general taxonomy of human traffic interactions, such as \citep{markkula2020defining}, our taxonomy is a further development of implicit communication. We develop the taxonomy based on common determinants of traffic interactions, such as, who claims the right-of-way, whether an agent relinquishes that right, whether an agent responds to the actions of other agents, and other alternate ways of resolving conflicts. We also provide an example mapping of strategies from the form that an AV strategic planner generates to the taxonomy developed in this paper, followed by an experimental evaluation of two popular models of strategic behavior, QLk and SP$\epsilon$NE, based on one pedestrian-vehicle and one vehicle-vehicle interaction scenario in simulation with respect to the taxonomy. Finally, we also provide real-world interaction video clips to illustrate the usefulness of the taxonomy.
\section{Taxonomy of strategies}
\begin{figure}[t]
         \centering
         \includegraphics[width=0.8\linewidth]{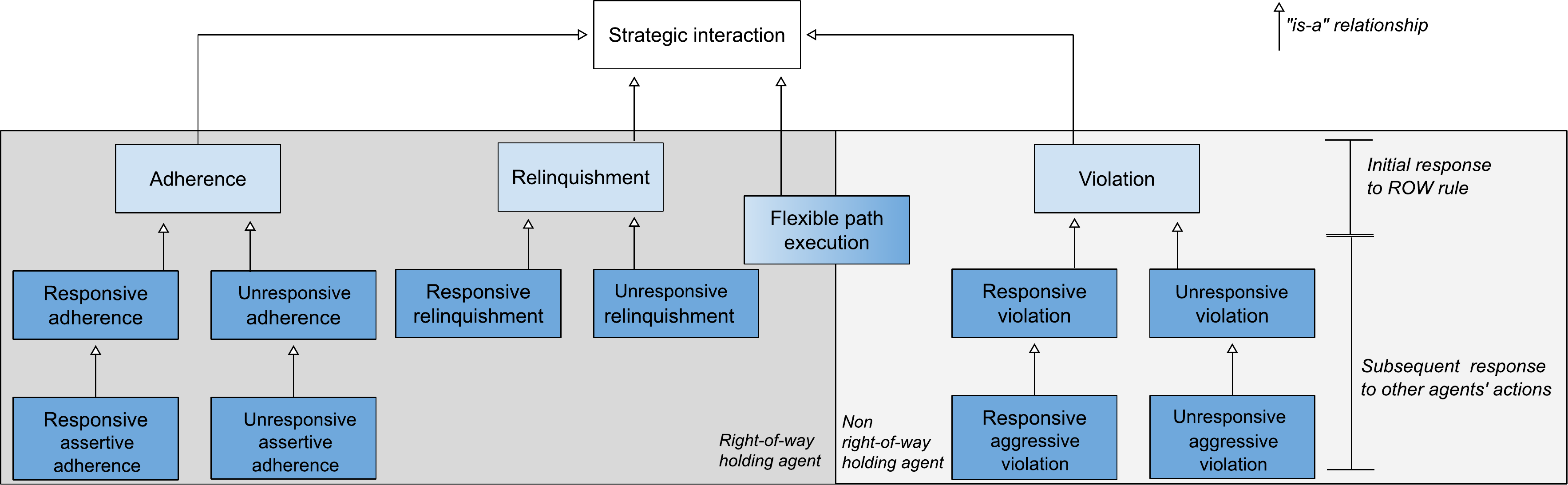}
         \caption{Schematic diagrams showing the relation among the taxonomy of strategies.}
         \label{fig:taxonomy_diag}
\end{figure}
The taxonomy we develop is focused on traffic situations that have a static conflict point, such as intersections and roundabouts. We first present the dimensions based on which the taxonomy is organized followed by the taxonomy.\\
\noindent \textbf{Right-of-way (ROW) rules.} One of the basic tasks in traffic navigation is conflict resolution. Conflict points are locations in traffic network where multiple lanes intersect, merge, or intersect with a crosswalk, and road users traversing those lanes need to behave in a co-ordinated manner to reduce risks of collision \citep{parker1989traffic}. Traffic rules play a major role in resolving the conflicts, and provide guidelines of behavior for all the road users involved. The way traffic rules resolve conflicts is by ROW assignment to a road user (e.g. who has the priority to proceed at an un-signalized intersection), where a road user holding the ROW has priority and can proceed to be the first to cross the conflict zone. In some jurisdictions, such as Austria, traffic rules even require road users relinquishing their ROW to indicate that through a signal. Therefore, the first dimension of the taxonomy is based on how road users (agents) behave in relation to the ROW rule in a traffic conflict.\\
\noindent \textbf{Responsiveness.} The second dimension of the taxonomy is based on how agents behave in relation to other agents' actions. A basic assumption of game-theoretic modeling of traffic is that the agents play a common game, where the each agent is aware of the other agents in the game. On the other hand, aspects such as distraction, mis-attention, and occlusion, lead to circumstances where one agent may not be aware of the other agent; and therefore, due to these arguably natural aspects of traffic and human behavior, the common game assumption breaks down. For humans, checking whether another road user is aware of them and at the same time making the other road user aware of their presence is a skill that we learn over time and takes the form a various non-verbal modalities of communication \citep{rasouli2017agreeing}. One such modality is through kinematic motion patterns \citep{dey2017pedestrian, aladawy2019eye}, where, based on changes in motion trajectories (e.g. pedestrians slowing down for a turning vehicle), a road user may communicate an acknowledgement that they are aware of another road user. Therefore, the second dimension of the taxonomy, \emph{responsiveness}, is built upon the above idea where we denote an agent's strategy to be responsive if it involves changes in the motion characteristics in response to the actions of other agents. Whereas following the default ROW rule can resolve the game in one round, the second dimension becomes salient when, due to miscommunication or agents not following the default rule, the game goes to subsequent rounds.\\
Fig. \ref{fig:taxonomy_diag} shows the taxonomy of strategies categorized based on the two dimensions of initial response to ROW rule (lighter boxes) and subsequent response to other agents' actions (darker boxes). Anonymized links to real-world snippets showing examples for each category is noted in Table 2 in the Appendix.
\label{sec:taxonomy}
\subsection{Taxonomy}
\subsubsection{Adherence}
Adherence is a class of strategy where an agent holding the ROW starts to proceed. Depending on the subsequent state of the game, and whether there is a change in characteristics of their trajectory, an adherence strategy can be further categorized as responsive or unresponsive.\\
\begin{figure}[t]
         \centering
         \includegraphics[width=0.8\linewidth]{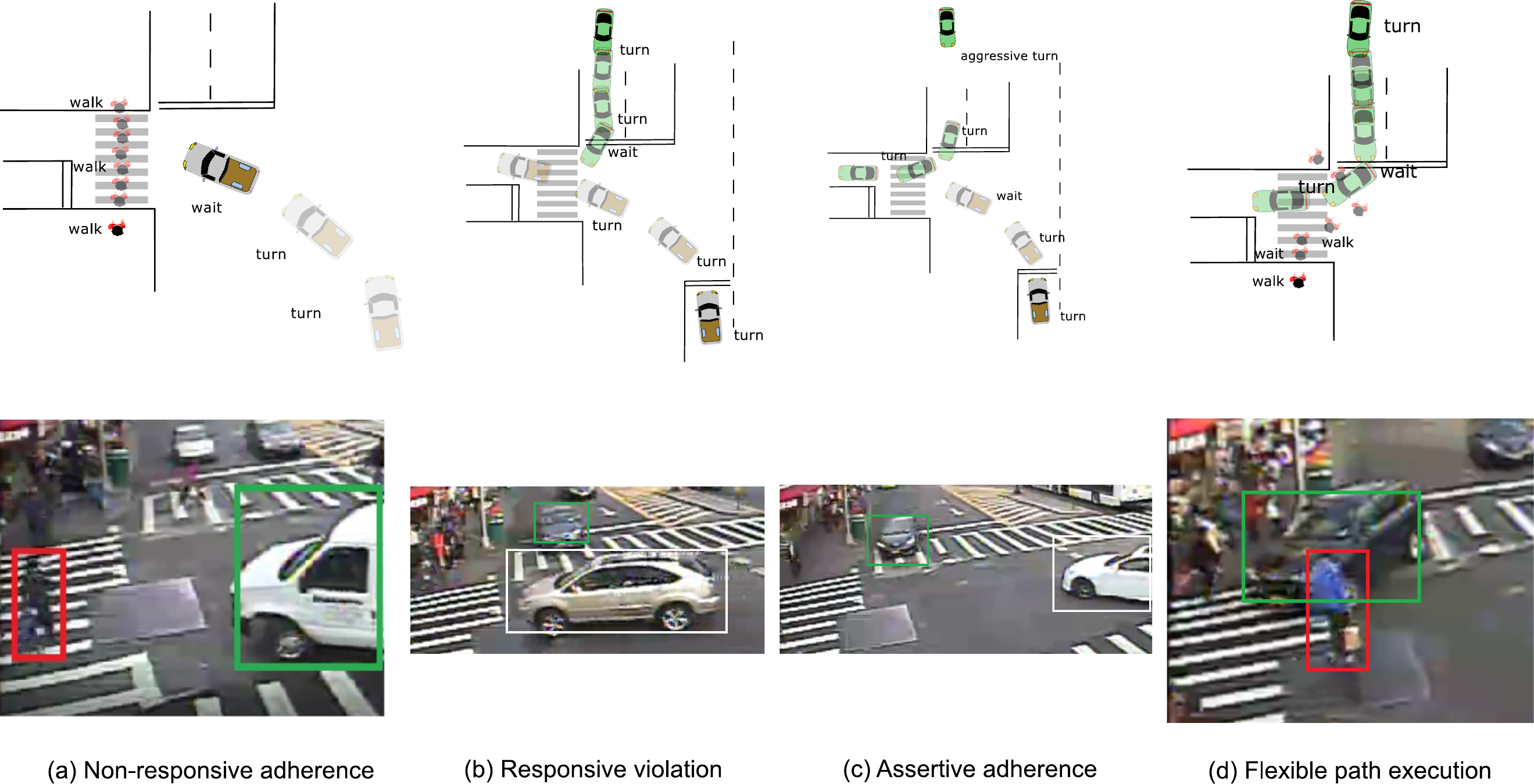}
         \caption{}
         \label{fig:strat_123}
\end{figure}
\noindent \textbf{\textit{Unresponsive adherence }(UA)}. In an unresponsive adherence strategy, an agent claims their ROW by starting to proceed, and irrespective of the actions of other agents' in the vicinity with whom they are in conflict with, they do not change their motion trajectory. A typical unresponsive adherence strategy is moving with a steady velocity even in the face of other conflicting agents' (who do not have the ROW) attempts to violate that right. Fig. \ref{fig:strat_123}a shows an example of unresponsive adherence. In this scenario, a pedestrian at a crosswalk has the ROW and starts crossing the crosswalk, while a white left turning vehicle in conflict (who does not have the ROW) starts to proceed with the left turn at the same time. An UA strategy on the part of the pedestrian is to continue walking at a steady speed, as opposed to slowing down in response to the white vehicle's action.\\
Although the UA strategy is motivated along the lines of whether or not a road user is aware of another conflicting road user, there can also be alternate explanations for this type of strategy. For example, while it is possible that in the above example, the pedestrian did not notice the white vehicle moving into the crosswalk, it is also possible that the pedestrian kept moving at a steady speed because they anticipated the white vehicle will slow down and wait after noticing them to be unresponsive, in other words, UA just being the optimal strategy from the pedestrian's perspective. \par
\noindent \textbf{\textit{Responsive adherence} (RA)}. Similar to UA, responsive adherence also implies that a ROW holding road user claims that right by starting to proceed. However, if a non ROW holding conflicting agent attempts to violate that right and proceeds at the same time, the ROW holding agent may demonstrate a change in their trajectory, for example, by slowing down and proceeding cautiously. In the example of Fig. \ref{fig:strat_123}a, a responsive adherence strategy on the part of the pedestrian can be a) slowing down and proceeding with caution upon observing the turning white vehicle, b) increasing their speed to move through the crosswalk fast to clear the way for the turning vehicle, and c) slowing down and waiting for the turning white vehicle to pass. Although these individual strategies represent different ways of dealing with the conflict, they demonstrate commonality in terms of the pedestrian being aware of the vehicle, a signal being established that they are playing a common game by their reaction to the white vehicle, as well as the initial willingness of the pedestrian to claim their ROW status.\\
\noindent \textbf{\textit{Responsive and unresponsive assertive adherence} (RAA and UAA)}. Assertive adherence is a specific type of adherence strategy, where a road user holding the ROW starts to proceed aggressively (aggression can be indicated by a higher than normal jerk and acceleration) in order to dissuade any conflicting road user from violating their right. The goal of this strategy is to signal a strong willingness to defend the ROW. Depending on whether the agent subsequently modifies their trajectory characteristics as a response to other agents' action, an assertive adherence can be responsive or unresponsive. Fig. \ref{fig:strat_123}c shows a responsive assertive adherence scenario, where, the left turning vehicle starts to turn without holding the ROW; however, the right turning vehicle in this case starts an aggressive turn to dissuade the other vehicle from violating their ROW, but subsequently changes their maneuver to a non-aggressive turn when the white vehicle relents.\\
\subsubsection{Relinquishment}
Relinquishment strategy is the opposite of adherence, where an agent holding the ROW chooses to wait and let another agent (who does not hold the ROW at that moment) to proceed instead. \\ \noindent \textbf{\textit{Unresponsive relinquishment} (UR)}. In a unresponsive relinquishment strategy, the agent continues to wait for the other agent to pass even when the other agent keeps waiting. \\ 
\noindent \textbf{\textit{Responsive relinquishment} (RR)}. Under responsive relinquishment, an agent after initially relinquishing their ROW reclaims the right and proceeds if the other agent continues to wait.
\subsubsection{Violation}
As the name suggests, this type of strategy is demonstrated by an agent who does not hold the ROW in the game, but decides to proceed. Similar to adherence strategies, ROW violation strategies can also belong to two categories, responsive and unresponsive.\\
\noindent \textbf{\textit{Unresponsive violation} (UV)}. In this strategy, an agent without holding the ROW starts to proceed, and continues while not responding to the ROW holding agents' attempt to reclaim their right. Under such a scenario, to avoid a collision, some of the strategies that the other agent can generate are unresponsive relinquishment, and responsive adherence of the type where they slow down and wait. Responsive adherence of speeding up can also avoid a collision but may be much more riskier choice.\\
\noindent \textbf{\textit{Responsive violation} (RV)}. In this type of ROW violation, a vehicle without holding the ROW starts to proceed; however, as a response to another ROW holding agent reclaiming their right, they can respond by changing their trajectory during the course of the game. Fig. \ref{fig:strat_123}b illustrates an example of responsive violation on the part of the white left-turning vehicle, which on account of not holding the ROW should have waited for the right turning vehicle. However, after starting to execute the turn, upon observing the right turning vehicle (who had the ROW) not relinquishing their right, speeds up to complete the turn fast in order to avoid getting stuck in middle of the intersection. The white left turning vehicle in the previous example of Fig. \ref{fig:strat_123}c also demonstrates responsive violation of a different flavor, where it first starts to proceed without holding the ROW, but eventually relents when the right turning vehicle demonstrates an assertive adherence strategy.\\
\noindent \textbf{\textit{Responsive and unresponsive aggressive violation} (RAV and UAV)}. Similar to assertive adherence, if a violating vehicle starts to proceed aggressively, such a strategy is categorized as aggressive violation (the term aggressive is commonly used as a driving behavior indicator \citep{sagberg2015review} and in this case is used instead of the word assertive to indicate that the agent did not hold ROW). Depending on whether an agent demonstrating aggressive violation later changes its trajectory characteristics based on the response of the other agents' action, the strategy can be of responsive or unresponsive flavor. 
\subsubsection{Flexible path execution (FP)}
The strategies presented until now are primarily based on a road user's choice of trajectory. Without the temporal component of a trajectory, just the sequence of locations (called a path) is generally determined by navigable traffic regions, such as vehicle lanes for vehicles, bike lanes for bicycles, and sidewalks and crosswalks for pedestrian. Common strategies in traffic are variations over the trajectories by agents changing their movement velocities, while changes in path are often minimal and are kept within the prescribed navigable regions. However, in some cases, road users may choose to execute an alternate path as a way to resolve a conflict. This alternate path execution strategy may be executed by an agent regardless of their ROW status, and is demonstrated by agents choosing an alternate path to their desired goal location. Fig. \ref{fig:strat_123}d shows an example of an alternate path execution strategy that follows after a (RA,RV) strategy. In this scenario, a pedestrian having the ROW starts to proceed over the crosswalk, and at the same time a right turning vehicle continues to turn through the intersection thereby attempting to violate the ROW. Both agents generate responsive strategies by observing each others actions, and chooses to wait for the other agent to cross as their next action, thereby leading to a deadlock. In order to resolve that deadlock, instead of reclaiming their ROW over the crosswalk, the pedestrian chooses to go around the back of the vehicle to the other side of the crosswalk. This type of strategy can be an initial action in the game as well as subsequent response and can be demonstrated by both ROW and non ROW holding agent (noted by the placement of the block in Fig. \ref{fig:taxonomy_diag}). \par
\subsection{Intermediate outcome --- deadlock.}
Although not a strategy in and of itself, deadlocks are momentary states in the game where agents in conflict stop and wait in order to avoid a collision. Deadlocks follows from certain strategy choices (for example, the case of (RA,RV) in the previous example), and need a way of resolution in the subsequent steps of the game. It can also result from a combination of relinquishment and adherence strategies where both agents wait for the other to proceed. Attempted resolutions of a deadlock can take one of the following forms --- a ROW holding agent proceeds (RA), a non ROW holding agent proceeds (RV), both proceeding at the same time (RA,RV), and any of the agents involved in the deadlock chooses an alternate path to their destination (FP).
\section{Mapping strategies to taxonomy}
\begin{table}[b]
\centering
\small
\begin{tabular}{lcc}
\toprule
ROW status & Maneuver strategy regex & Matched taxonomy\\
    \midrule
    \multirow{6}{*}{For agent holding ROW}&$w^{+}$&unresponsive relinquishment (UR)\\
    &$p^{+}$&unresponsive adherence (UA)\\&$p_{a}^{+}$&unresponsive assertive adherence (UAA)\\&$w^{+}p\{w|p\}^{*}$&responsive relinquishment (RR)\\&$p^{+}w\{w|p\}^{*}$&responsive adherence (RA)\\&$p_{a}^{+}w\{w|p\}^{*}$&responsive assertive adherence (RAA)\\
    \midrule
    \multirow{6}{*}{For agent not holding ROW}&$w^{+}$&unresponsive adherence (UA)\\
    &$p^{+}w\{w|p\}^{*}$&responsive violation (RV)\\&$p_{a}^{+}w\{w|p\}^{*}$&responsive aggressive violation (RAV)\\&$p^{+}$&unresponsive violation (UV)\\&$p_{a}^{+}$&unresponsive aggressive violation (UAV)\\&$w^{+}p\{w|p\}^{*}$&responsive adherence (RA)\\
    \bottomrule
\end{tabular}
\normalsize
\caption{Regular expression mapping strategies to taxonomy}
\label{tab:strat_2_tax}
\end{table}
In order for the developed taxonomy to be usable in the evaluation of AV planners, there needs to be a well-defined translation of the strategies that a strategic planner generates to one of the strategy categories developed in the taxonomy. While developing a translation for every strategic planner proposed in the literature is beyond the scope of this paper, in this section we illustrate one example. Hierarchical game theoretic planners, similar to the ones proposed in \citep{sarkar2021solution}, generates strategies that are a combination of high-level maneuver and a low-level trajectory segment, where, each trajectory segment at a node in the game belongs to one of the two types of high-level maneuver, \emph{wait} or \emph{proceed}. We denote each maneuver as a symbol, where $w$ corresponds to \emph{wait} maneuver, $p$ corresponds to \emph{proceed} maneuver, and $p_{a}$ is an \emph{aggressive} proceed maneuver defined based on the maximum acceleration of the trajectory. The maneuver choices in the strategy can then be expressed as a regular expression (regex), where each literal in the regex corresponds to a maneuver choice at each node of the game starting from $t=0$ (Table \ref{tab:strat_2_tax}). To illustrate few examples, for an agent holding the ROW, a unresponsive relinquishment strategy involves generation of \emph{wait} maneuver regardless of the actions of the other agents, which can be expressed as the regex $w^{+}$. Similarly, a regex $w^{+}p\{w|p\}^{*}$ expresses strategies where the agent chooses wait maneuver in the first node, followed by at least one proceed choice at any node along the game. For a non-ROW holding agent, this would imply responsive adherence since they adhere to the ROW rule by waiting; however, they can subsequently respond to the other agent's action by choosing a proceed maneuver. On the other hand, for a ROW holding agent, the same regex is of responsive relinquishment type. While we defined the process of matching a strategy to one of the taxonomy purely based on the maneuver choices, one can also think of a more granular process of matching the strategies based on the trajectory choices; for example, based on bounds of the acceleration to define what counts as a steady movement for unresponsive strategies.

\section{Evaluation of strategic behavior models}
One of the goals of developing a taxonomy is to establish a common language based on which different models of strategic behavior can be compared and evaluated. For the context of this paper, we perform an evaluation of two popular classes of models proposed for the problem of strategic behavior planning in AV literature --- QLk model \citep{tian2018adaptive,li2019decision} and Nash equilibrium \citep{pruekprasert2019decision,Geiger_Straehle_2021,michieli2018game}, and discuss the type of strategies that each model generates. More specifically, for the Nash equilibrium based model, we select subgame perfect $\epsilon-$Nash equilibrium (SP$\epsilon$NE) \citep{flesch2016refinements} as the solution concept for even comparison with QLk in regards to the support for bounded rational agents in a dynamic game setting. For the Qlk model, there is a choice to be made on the solution concept that a level-0 agent uses based on the caveat that level-0 behavior should be non-strategic. To that end, we select a maxmax model due to its higher alignment with naturalistic driving behavior compared to other non-strategic models \citep{sarkar2021solution}. In our implementation of the QLk model, all agents are modeled as level-1, i.e., internally they model other agents as level-0 \citep{wright2010beyond}.
\subsection{Game structure}
\label{sec:game_tree}
\begin{wrapfigure}{r}{0.5\textwidth}
  \centering
    \includegraphics[width=.5\linewidth]{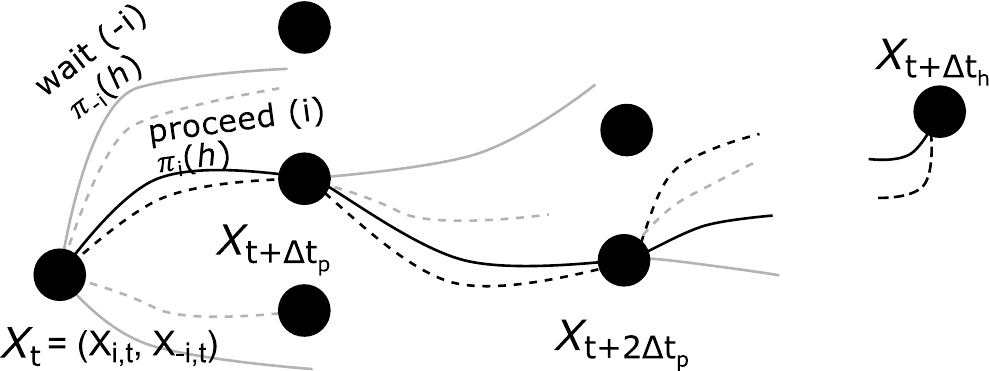}
    \caption{Schematic representation of the dynamic game. Each node is embedded in a spatio-temporal lattice and nodes are connected with a cubic spline trajectory.}
  \label{fig:game_tree}
\end{wrapfigure}
The game is constructed as a dynamic game between two road users in conflict modelled as a sequence of simultaneous move games played every $\Delta t_{p}=1.3$ secs. starting at time $t=0$ over a horizon of $\Delta t_{h}=4$ secs. Detailed game tree construction is included in the Appendix. Strategies in the game are pure strategies representing a trajectory choice and expressed in the behavior form, $\pi_{i}(h_{t}) \in \mathrm{T_{i}(X_{t})}$, where $\mathrm{T_{i}(X_{t})}$ is the set of valid trajectories (based on the kinematics limits of the vehicles and pedestrians) that can be generated by agent $i$ at the node $X_{t}$, and $h_{t}$ is the history of the game or the path from from the root node till $X_{t}$. The utilities in the game are multi-objective utilities consisting of two components --- \textbf{safety} $u_{s,i}(\pi_{i}(h),\pi_{-i}(h)) \in [-1,1]$ (modelled as a sigmoidal function that maps the minimum distance gap between trajectories to a utility interval [-1,1]) and \textbf{progress} $u_{p,i}(\pi_{i}(h),\pi_{-i}(h)) \in [0,1]; \forall i, -i$ (modelled as a linear function that maps the trajectory length in meters to a utility interval [0,1]). Each type ($\gamma_{i} \in [-1,1]$) of agent combines the two objectives based on a lexicographic threshold \citep{LiChangjian19} derived from their own type value $\gamma_{i}$, where the combined utility $u_{i}(\pi_{i}(h),\pi_{-i}(h))$ is equal to $u_{s,i}(\pi_{i}(h),\pi_{-i}(h))$, i.e., the safety utility when $u_{s,i}(\pi_{i}(h),\pi_{-i}(h)) \leqslant -\gamma_{i}$, and otherwise $u_{i}(\pi_{i}(h),\pi_{-i}(h)) = u_{p,i}(\pi_{i}(h),\pi_{-i}(h))$, i.e., the progress utility. A higher value of $\gamma_{i}$ indicates higher risk tolerance since an agent with type $\gamma_{i}$ takes into account the safety utility only in situations where the value is less than equal to $-\gamma_{i}$. In our experiments, we discretize the types by increments of 0.5. The utilities at a node with associated history $h_{t}$ are calculated as discounted sum of utilities over the horizon of the game conditioned on the strategy profile $\sigma$, type $\gamma_{i}$, and discount factor $\delta (=0.5)$ as follows
\begin{equation*}
\small
    \sum\limits_{k=1}^{\floor*{\Delta t_{h}-t \backslash \Delta t_{p} }} \delta^{k}u_{i}(\pi_{i}(h_{t+k-1}),\pi_{-i}(h_{t+k-1});\sigma,\gamma_{i}) + \mathcal{N}u_{i,C}
\normalsize
\end{equation*}
where $u_{i,C}$ is the terminal utility estimated based on agents continuing on the trajectory chosen at the last decision node of the game tree for another $\Delta t_{h}$ seconds, and $\mathcal{N}$ is a normalization constant. In order to dissuade ROW violation, an agent is penalized $\tau (=0.25)$ in their safety utility value for not following ROW rule.

\subsection{Simulation runs.}

\begin{figure}[t]
  \centering
  \begin{subfigure}[b]{0.3\textwidth}
  \centering
  \includegraphics[width=0.75\textwidth]{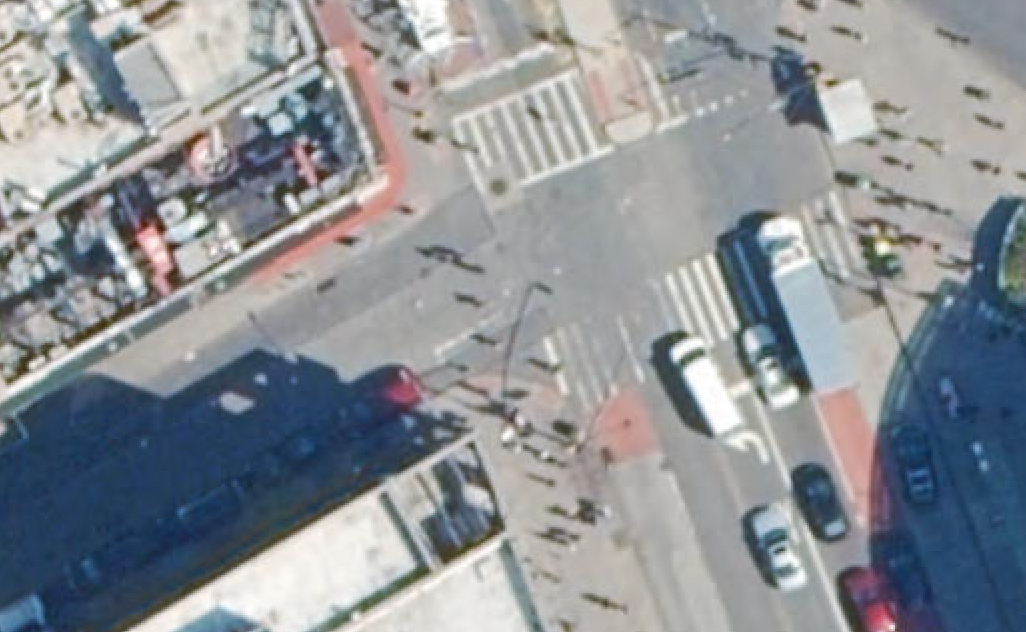}
  \caption{}
  \end{subfigure}
  \begin{subfigure}[b]{0.3\textwidth}
    \centering
    \includegraphics[width=0.5\textwidth]{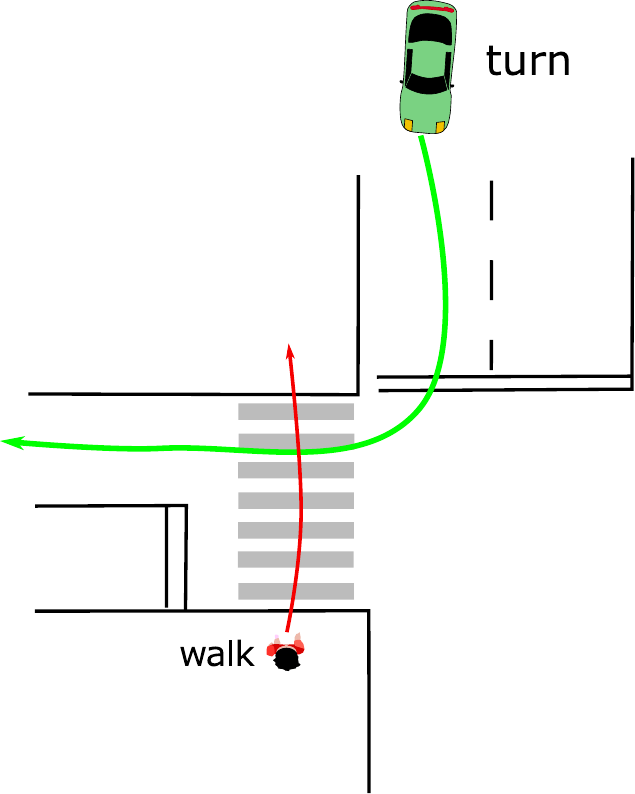}
    \caption{}
    \end{subfigure}
    \begin{subfigure}[b]{0.3\textwidth}
    \centering
    \includegraphics[width=0.6\textwidth]{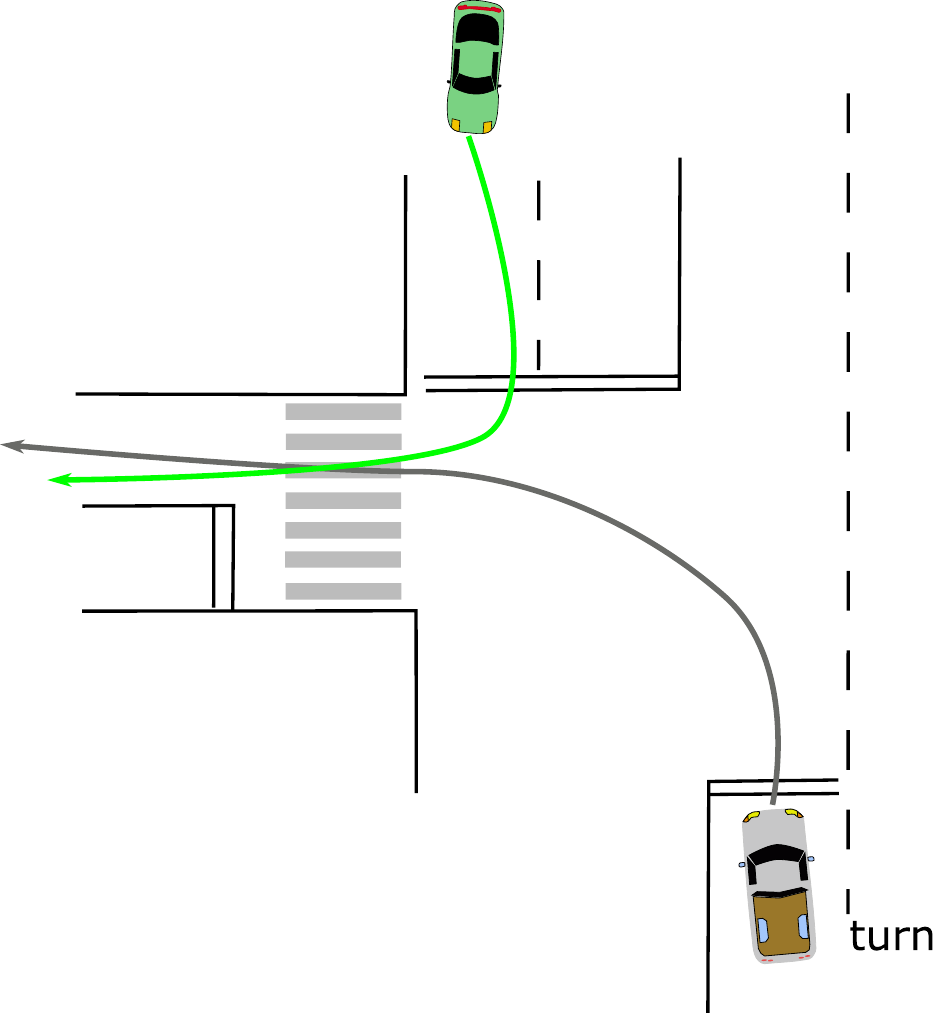}
    \caption{}
    \end{subfigure}
    \caption{Pedestrian-vehicle and vehicle-vehicle interaction scenario}
  \label{fig:simul_scenes}
\end{figure}

For each of the models, we run simulations over two scenarios, one pedestrian-vehicle and one vehicle-vehicle interaction scenario. For the pedestrian-vehicle scenario (Fig. \ref{fig:simul_scenes}b), a pedestrian (who has the ROW) has just started to cross the crosswalk at the moment of initiation of the game, and a right turning vehicle should ideally wait for the pedestrian to cross. For the vehicle-vehicle interaction scenario (Fig. \ref{fig:simul_scenes}c), a right turning vehicle holding the ROW has just started to execute its turn, and the left turning vehicle should wait for the right turning vehicle to cross before executing the turn. Both scenarios are simulated based on a map of a New York City intersection (Fig. \ref{fig:simul_scenes}a). The scenarios are initiated based on different initial velocities of the agents (1.3 to 1.8 $\text{ms}^{-1}$ for pedestrians and 1 to 12 $\text{ms}^{-1}$ for vehicles) as well as all combinations of agent types $\gamma \in$ [-1,1] with 0.5 increments. In total, there are a total of 1250 games for the pedestrian-vehicle scenario, and 2500 games for the vehicle-vehicle scenario. All the games are solved in a complete information setting with regards to agent types, the parameter $\epsilon$ for SP$\epsilon$NE model is 0.1, and precision parameter ($\lambda$) for QLk is 1. \\
\begin{figure}[t]
  \centering
    \includegraphics[width=\linewidth]{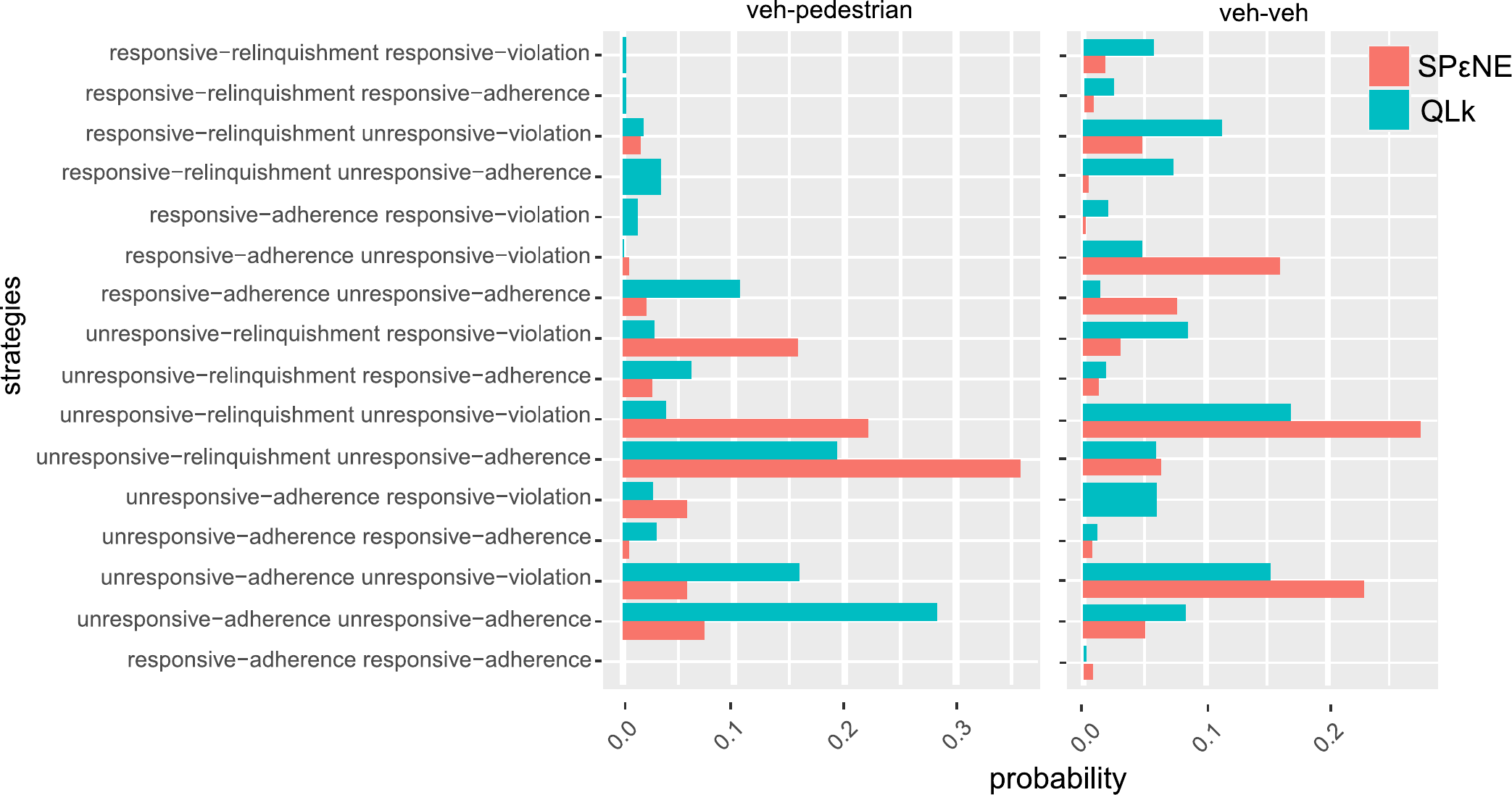}
    \caption{Distribution of strategies for SP$\epsilon$NE and QLk model for the two simulation scenarios.}
  \label{fig:all_strat_tax}
\end{figure}
\subsection{Results}
Fig. \ref{fig:all_strat_tax} shows the distribution of the basic strategies (without the aggressive and assertive modifiers) corresponding to each behavior model. We see that there is a wide range of strategies that each model generates based on the initiating situations of the game. However, for agents holding the ROW, unresponsive strategies (both relinquishment and adherence) are the more common type of strategies generated by both models. Given that the simulation did not incorporate an explicit model of distraction or mis-attention, the results demonstrate that the unresponsive strategies can also be the  optimal strategies compared to the other alternatives in some situations.\\
Certain strategies are frequently seen in both models, for example, (UR,UA), i.e., both pedestrian and vehicle waiting for each other in vehicle-pedestrian interaction, and (UR,UV), i.e., pedestrian waits and the vehicle executes the turn for vehicle-vehicle interaction. Whereas, there are other strategies in which the models show considerable disagreement. For vehicle-pedestrian scenario, the most common strategy in the QLk model is unresponsive adherence for both agents (the vehicle waiting for the pedestrian to cross), and is observed in approximately 30\% of the games. Although this is arguably the ideal strategy in this scenario, it is not a commonly observed strategy in SP$\epsilon$NE model (less that 10\% of the games). Similar disagreement is seen for vehicle-vehicle scenario with respect to (RA,UV) strategy, i.e., the right turning vehicle claims their ROW and starts to move but waits on account of the left turning vehicle violating the ROW in a unresponsive way; this strategy is more commonly observed in SP$\epsilon$NE model than QLk. \\
\begin{figure}[t]
  \centering
    \includegraphics[width=\linewidth]{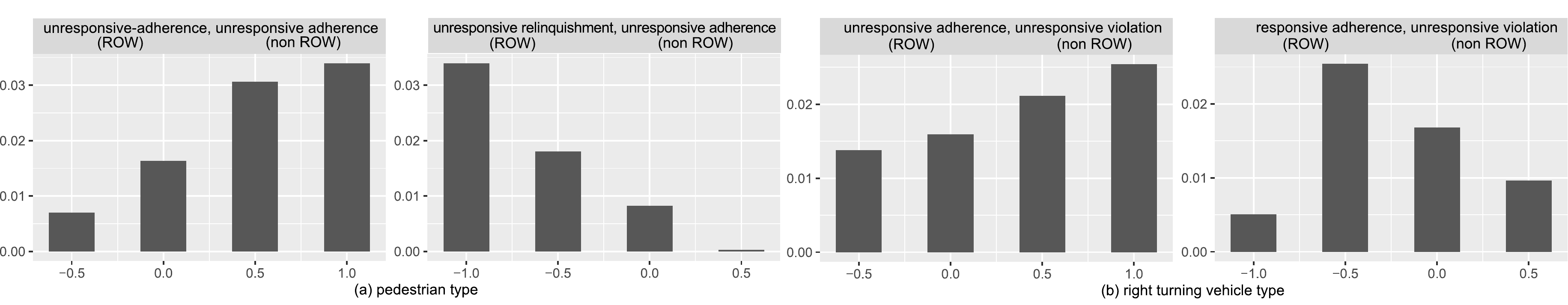}
    \caption{Distribution of agent types for ROW holding agent for frequently generated strategies.}
  \label{fig:type_distr}
\end{figure}
The generated strategies also depend on the type value of the agents (recall that higher the type value of an agent, higher is their risk tolerance). Figure \ref{fig:type_distr}a shows the distribution of the two most common strategies observed in the QLk model for the pedestrian-vehicle interaction scenario with respect to the pedestrian type value. Since with higher types, the chances of the pedestrian generating a proceed action becomes higher, we see in the two plots of Figure \ref{fig:type_distr}a, that with higher types the probability of unresponsive adherence is higher and unresponsive relinquishment is lower.\\
There is also an association between the agent type value and whether the strategies they generate are responsive or unresponsive. Fig. \ref{fig:type_distr}b contrasts (UA,UV) and (RA,UV) strategies, which differ only on responsiveness. The first plot of Fig. \ref{fig:type_distr}b shows what we would normally expect; the chances of a vehicle claiming their ROW even in the face of the right being violated will increase with their type, reflected in the increasing probability of the unresponsive adherence strategy. However, as we see from the second plot, the probability of responsive adherence strategy peaks at medium risk tolerance (-0.5). This is because if the risk tolerance is too low, then the ROW holding agent will not claim their ROW at all and thereby not generate an adherence strategy. On the other hand, if their risk tolerance is too high, then they will demonstrate a unresponsive adherence rather than an responsive one.\\ 
Based on the above analysis, first, given that the two popular models generate a wide range of strategies based on different game situations, an AV will need to carefully evaluate whether a chosen strategic model is appropriate based on the type of strategy they generate in that specific situation. Second, since certain type of strategies are more commonly generated by agents with specific types, an AV can use the observed strategies to form a possible hypothesis about a road user, for example, if a road user generates a responsive strategy, then their risk tolerance may be lower than someone generating an unresponsive strategy, and an AV needs to take into account that information.
\section{Conclusion}
We presented a taxonomy of human strategic interactions in traffic conflicts that is built upon the basic dimensions of how agents response to their right-of-way as well as to each others' actions in a game. We demonstrated a way to map strategies from a form generated by a typical strategic planner to the ones in the taxonomy. Based on our evaluation of two popular solution concepts used in strategic planning, we highlighted the relation between the types of strategies generated to the agents' risk tolerance. On top of the presented taxonomy, one can build a formal framework of the emergent communication as well as outcomes based on the different combination of strategies, e.g., resolution based on traffic rules, deadlock, severe conflict (near-miss, crash, etc.), successful violation, etc. We hope that along with the development of a more formal framework for strategic safety specifications over the taxonomy as a future work, the taxonomy can also be extended for other traffic situations with discretionary actions and dynamic conflict points, such as lane changes, which were not covered in this paper.     
\section*{Acknowledgement}
We thank New York City Department of Transport (DoT) for the videos. The videos are provided to the authors were in low resolution in order to maintain the privacy and anonymity of the road users.
\bibliographystyle{plainnat}
\bibliography{main}

\appendix

\section{Appendix}
\begin{table}[h]
\centering
\small
\begin{tabular}{p{2cm}p{8cm}p{2.75cm}}
\toprule
Category & Link to an example snippet & Description\\
    \midrule
    unresponsive adherence&\url{https://mega.nz/file/X8tUEboB#WmliGbu_m8X6Q9-f6YoJbV-1-V-7TCWTfC2971A4T7k} & Person walking on left crosswalk \\
    responsive adherence& \url{https://mega.nz/file/2tkkARqK#rTzYGriNF5boTyec29yV7ISCQ3ZKs1ZbfLY8CCcHy-A} &Person walking on left crosswalk\\
    responsive assertive adherence&\url{https://mega.nz/file/mktGHZTb#pxDaalA7DrkV2IspNPUqtaHtJLB_2-x0WxG_ORgXufw}&Right turning vehicle\\
    unresponsive relinquishment& \url{https://mega.nz/file/yps2hbYD#NCLAXp9seDZl5BfBl80wgQfLkhU4X4kQTNjhrOcPUFo}&Person standing on left crosswalk\\
    unresponsive violation& \url{https://mega.nz/file/ThlUUZpZ#O1n_RJZ7unKJFhKtqRafZswDk_5GjYgfTSFG7T9sGM0}&Person walking on bottom-right crosswalk\\
    responsive violation&\url{https://mega.nz/file/n90EhZgQ#Ck55tOTRqJg_baa2B6eM3Y-fXwvp0sJhiyMvB0hv5xM} &silver left-turning vehicle\\
    flexible path execution&\url{https://mega.nz/file/XpsAxTJT#Nmw_VyXO-lqadl-tfVnabhlXXPzXVlJ8fiGiGzFVqYc} &Person walking on left crosswalk \\
    deadlock& \url{https://mega.nz/file/nskGERgT#76HIL3v7RAnF4GK2aaPUOn6lgVa8wB3N1T9quaVDfeI}&Between person walking on left crosswalk and right-turning vehicle \\
    \bottomrule
\end{tabular}
\normalsize
\caption{Links to real-world examples of each category of strategy developed in the taxonomy}
\label{tab:appendix_tab}
\end{table}
\subsection{Game tree construction details.}
The nodes ($X_{t}$) of the game tree are the joint system states of each agent $i \in N$ in the game, where an agent's state is represented as a vector $X_{i,t} = [x,y,v_{x},v_{y},\dot{v_x},\dot{v_y},\theta]$. $x,y$ are positional co-ordinates on $R^{2}$, $v_{x},v_{y}$ are lateral and longitudinal velocity in the body frame, $\dot{v_x},\dot{v_y}$ are the acceleration components, and $\theta$ is the yaw of the trajectory. For vehicles, the actions are cubic spline trajectories generated based on kinematic limits of vehicles with respect to bounds on lateral and longitudinal velocity, acceleration, and jerk. For pedestrians, the action trajectories are generated based on a constant velocity model with different ranges of walking speed. At each node, the trajectories are generated with respect to high-level maneuvers, namely, \emph{wait} and \emph{proceed} maneuvers. For wait maneuver trajectories, a moving agent decelerates and comes to a stop, and for proceed maneuvers, an agent maintains its moving velocity or accelerates to a range of target speeds.\\

\subsection{Code and data}
All code and data used in the paper is available at this link \url{https://mega.nz/file/vpshyCaD#tnF5kzxe7TAG5wmNOM1GErmrxD2SHTC0p42gh9mM7OU}. Follow instructions included in Instructions.txt to reproduce the experiments.


\section*{Checklist}

\begin{enumerate}

\item For all authors...
\begin{enumerate}
  \item Do the main claims made in the abstract and introduction accurately reflect the paper's contributions and scope?
    \answerYes{}`
  \item Did you describe the limitations of your work?
    \answerYes{The scope of the taxonomy for static conflict situations.}
  \item Did you discuss any potential negative societal impacts of your work?
    \answerNA{The paper develops a framework in the step towards safer autonomous vehicle development}
  \item Have you read the ethics review guidelines and ensured that your paper conforms to them?
    \answerYes{}
\end{enumerate}

\item If you are including theoretical results...
\begin{enumerate}
  \item Did you state the full set of assumptions of all theoretical results?
    \answerNA{}
	\item Did you include complete proofs of all theoretical results?
    \answerNA{}
\end{enumerate}

\item If you ran experiments...
\begin{enumerate}
  \item Did you include the code, data, and instructions needed to reproduce the main experimental results (either in the supplemental material or as a URL)?
    \answerYes{Details in the Appendix.}
  \item Did you specify all the training details (e.g., data splits, hyperparameters, how they were chosen)?
    \answerNA{}
	\item Did you report error bars (e.g., with respect to the random seed after running experiments multiple times)?
    \answerNA{}
	\item Did you include the total amount of compute and the type of resources used (e.g., type of GPUs, internal cluster, or cloud provider)?
    \answerYes{}
\end{enumerate}

\item If you are using existing assets (e.g., code, data, models) or curating/releasing new assets...
\begin{enumerate}
  \item If your work uses existing assets, did you cite the creators?
    \answerNA{}
  \item Did you mention the license of the assets?
    \answerNA{}
  \item Did you include any new assets either in the supplemental material or as a URL?
    \answerYes{}
  \item Did you discuss whether and how consent was obtained from people whose data you're using/curating?
    \answerNA{}
  \item Did you discuss whether the data you are using/curating contains personally identifiable information or offensive content?
    \answerYes{}
\end{enumerate}

\item If you used crowdsourcing or conducted research with human subjects...
\begin{enumerate}
  \item Did you include the full text of instructions given to participants and screenshots, if applicable?
    \answerNA{}
  \item Did you describe any potential participant risks, with links to Institutional Review Board (IRB) approvals, if applicable?
    \answerNA{}
  \item Did you include the estimated hourly wage paid to participants and the total amount spent on participant compensation?
    \answerNA{}
\end{enumerate}

\end{enumerate}


\end{document}